\documentclass[letterpaper, 10 pt, conference]{ieeeconf}  % Comment this line out if you need a4paper
\IEEEoverridecommandlockouts                              % This command is only needed if
\usepackage{graphics}    % for pdf, bitmapped graphics files
\usepackage{times}       % assumes new font selection scheme installed
\usepackage{amsmath}     % assumes amsmath package installed
\usepackage{amssymb}     % assumes amsmath package installed
\usepackage{graphicx}
\usepackage{algorithm}
\usepackage[noend]{algpseudocode}
\usepackage{booktabs}
\usepackage{siunitx}
\usepackage{mathtools}
\usepackage{hyperref}

\usepackage{subfigure}
\usepackage{color} % for commenting

\usepackage{relsize}

\def\secref#1{Sec.~\ref{#1}}
\def\figref#1{Fig.~\ref{#1}}
\def\tabref#1{Tab.~\ref{#1}}
\def\eqref#1{Eq.~(\ref{#1})}

\renewcommand{\vec}[1]{\mathbf{#1}}	%% vectors are bold
\newcommand{\mat}[1]{\mathbf{#1}}  	%% matricies are also bold
\newcommand{\set}[1]{\mathcal{#1}} 	%% sets are denoted by calligraphic letters

\newcommand{\RR}{\mathbb{R}} %% real numbers
\newcommand\Tstrut{\rule{0pt}{2.6ex}}         % = `top' strut
\makeatletter

\usepackage{xspace}
\DeclareRobustCommand\onedot{\futurelet\@let@token\@onedot}
\def\@onedot{\ifx\@let@token.\else.\null\fi\xspace}

\def\eg{e.g\onedot} 
\def\ie{i.e\onedot}

\def\etal{\emph{et al}\onedot}
\makeatother

\usepackage{array}
\newcolumntype{L}[1]{>{\raggedright\let\newline\\\arraybackslash\hspace{0pt}}m{#1}}
\newcolumntype{C}[1]{>{\centering\let\newline\\\arraybackslash\hspace{0pt}}m{#1}}
\newcolumntype{R}[1]{>{\raggedleft\let\newline\\\arraybackslash\hspace{0pt}}m{#1}}

\title{\LARGE \bf Learning an Overlap-based Observation Model \\ for 3D LiDAR Localization}

\author{Xieyuanli Chen  \and Thomas L\"abe \and 
Lorenzo Nardi   \and Jens Behley  \and Cyrill Stachniss\\
  \thanks{All authors are with the University of Bonn, Germany.}%
  \thanks{This work has been funded by the Deutsche Forschungsgemeinschaft (DFG, German Research Foundation) under Germany's Excellence Strategy, EXC-2070 - 390732324 - PhenoRob as well as under grant number BE 5996/1-1 and by the Chinese Scholarship Committee.
  }%
}

\begin{document}
\maketitle
\thispagestyle{empty}
\pagestyle{empty}

%%%%%%%%%%%%%%%%%%%%%%%%%%%%%%%%%%%%%%%%%%%%%%%%%%%%%%%%%%%%%%%%%%%%%%%%%%%%%%%%
\begin{abstract}
  Localization is a crucial capability for mobile robots and autonomous cars.
  In this paper, we address learning an observation model for Monte-Carlo localization using 3D LiDAR data. We propose a novel, neural network-based observation model that computes the expected overlap of two 3D LiDAR scans. 
  The model predicts the overlap and yaw angle offset between the current sensor reading and  virtual frames generated from a pre-built map. 
  We integrate this observation model into a Monte-Carlo localization framework and tested it on urban datasets collected with a car in different seasons.
  The experiments presented in this paper illustrate that our method can reliably localize a vehicle in typical urban environments. We furthermore provide comparisons to a beam-endpoint and a histogram-based method indicating a superior global localization performance of our method with fewer particles.
\end{abstract}

%%%%%%%%%%%%%%%%%%%%%%%%%%%%%%%%%%%%%%%%%%%%%%%%%%%%%%%%%%%%%%%%%%%%%%%%%%%%%%%%
\section{Introduction}
\label{sec:intro}

All mobile systems that navigate autonomously in a goal-directed manner need to know their position and orientation in the environment, typically with respect to a map. This task is often referred to as localization and can be challenging especially at the city-scale and in the presence of a lot of dynamic objects, \eg, vehicles and humans.
Over the last decades, a wide range of localization systems have been developed relying on different sensors.  Frequently used sensors are GPS receivers, inertial measurement units, cameras, and laser range scanners. All sensing modalities have their advantages and disadvantages. For example, GPS does not work indoors, cameras do not work well at night or under strong appearance changes, and LiDARs are active sensors and are still rather expensive.

\begin{figure}[t]
  \centering
  \vspace{0.25cm}
  \includegraphics[width=0.99\linewidth]{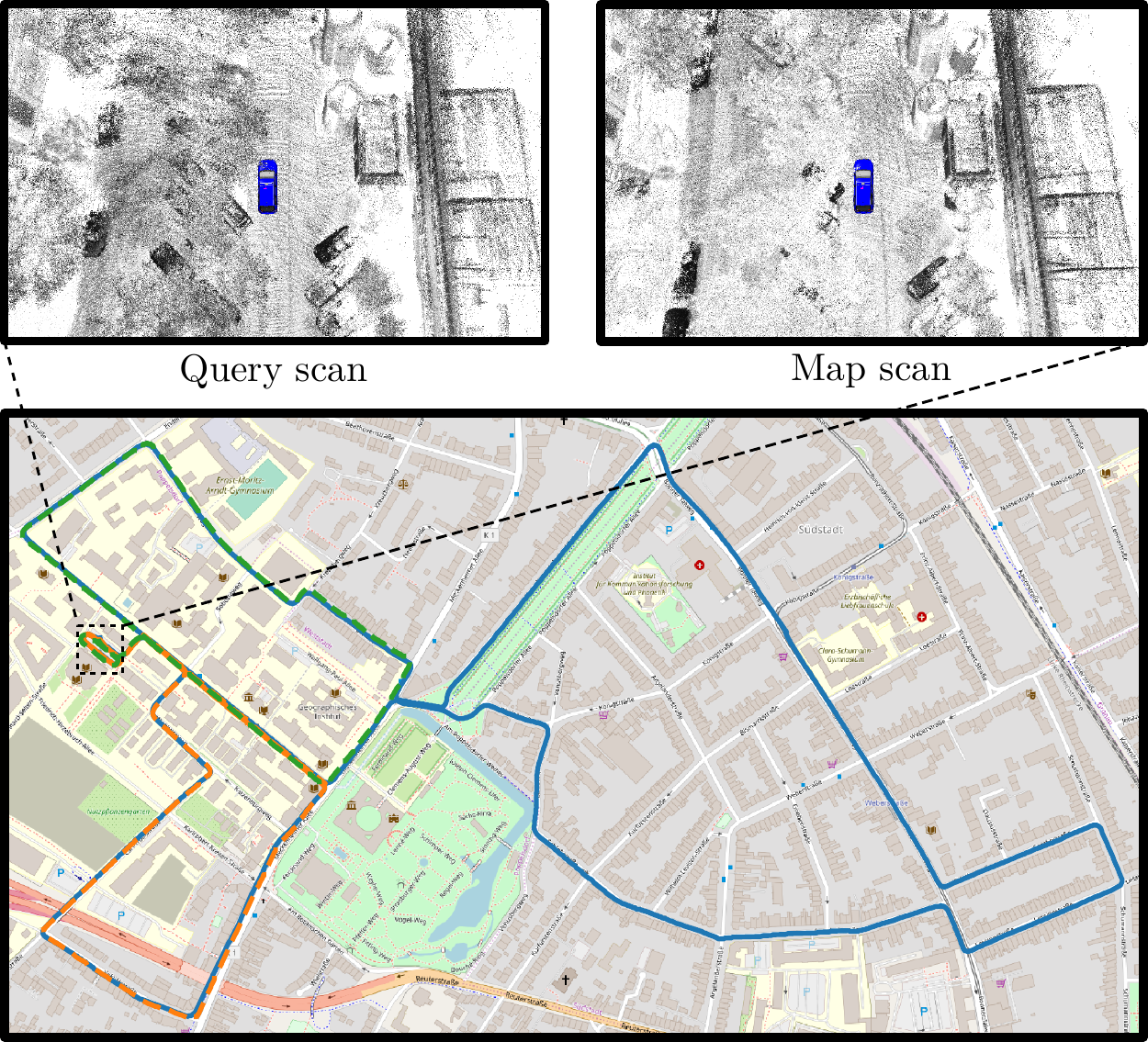}
  \caption{The image in the lower part shows the trajectories of the dataset used in this paper, overlayed on OpenStreetMap. 
  The blue trajectory represents the sequence used to generate a map for localization.
  The green and the orange trajectories represent two different test sequences. 
  The 3D point clouds shown in the upper part show LiDAR scans of the same place, once during mapping and once during localization.  Since the LiDAR data was collected in different seasons, the appearance of the environment changed quite significantly due to changes in the vegetation but also due to parked vehicles at different places. 
%  \vspace{-0.5cm}
  }
  \label{fig:motivation}
\end{figure}

Most autonomous robots as well as cars have a 3D LiDAR sensor onboard to perceive the scene and directly provide 3D data.  In this paper, we consider the problem of vehicle localization \emph{only} based on a 3D LiDAR sensor.
For localization, probabilistic state estimation techniques such as extended Kalman filters~(EKF)~\cite{cox1991tra}, particle filters~\cite{fox2001particle}, or incremental factor graph optimization approaches~\cite{smith1986ijrr} can be found in most localization systems today. Whereas EKFs and most optimization-based approaches track only a single mode, particle filters inherently support multi-modal distributions. Furthermore, PFs do not restrict the motion or observation model to follow a specific distribution such as a Gaussian. The motion model can often be defined quite easily based on the physics of vehicle motion. The observation model, however, is trickier to define. Often, these hand-designed models are used and they strongly impact the performance of the resulting localization system. Frequently used observation models for LiDARs are the beam-end point model, also called the likelihood field~\cite{thrun2005probrobbook}, the ray-casting model~\cite{dellaert1999icra}, or models based on handcrafted features~\cite{steder2011iros, zhang2018iros}. Recently, researchers also focused on learning such models completely from data~\cite{barsan2018corl, wei2019cvpr}.

In this paper, we address the problem of learning observation models for 3D LiDARs and propose a deep neural network-based approach for that. We explore the possibilities of learning an observation model based on OverlapNet~\cite{chen2020rss}  that predicts the overlap between two LiDAR scans, 
see~\figref{fig:motivation}, % add according to reviews.
where the overlap is defined as the ratio of points that can be seen 
from both LIDAR scans.  % add according to the reviews
In this work, we investigate this concept for 3D LiDAR-based observation models. 

%\clearpage
The main contribution of this paper is a novel observation model for 3D LiDAR-based localization. Our model is learned from range data using a deep neural network. It estimates the overlap and yaw angle offset between a query frame and map data. We use this information as the observation model in Monte-Carlo localization~(MCL) for updating the importance weights of the particles. Based on our novel observation model, our approach achieves online localization using 3D LiDAR scans over extended periods of time with a comparably small number of particles.

The source code of our approach is  available at:

\url{https://github.com/PRBonn/overlap_localization}

%%%%%%%%%%%%%%%%%%%%%%%%%%%%%%%%%%%%%%%%%%%%%%%%%%%%%%%%%%%%%%%%%%%%%%%%%%%%%%%%
\section{Related Work}
\label{sec:related}

Localization is a classical topic in robotics~\cite{thrun2005probrobbook}.
For localization given a map, one often distinguishes between pose tracking and global localization.
In pose tracking, the vehicle starts from a known pose and the pose is updated over time. 
In global localization, no pose prior is available. 
In this work, we address global localization using 3D laser scanners without assuming any pose prior from GPS or other sensors. Thus, we focus mainly on LiDAR-based approaches in this section. 

Traditional approaches to robot localization rely on probabilistic state estimation techniques. 
These approaches can still be found today in several localization systems~\cite{ma2019iros, chong2013icra}.
A popular framework is Monte-Carlo localization~\cite{dellaert1999icra,thrun2001ai,fox1999aaai}, which uses a particle filter to estimate the pose of the robot.

In the context of autonomous cars, there are many approaches building and using high-definition~(HD) maps for localization, i.e., tackling the simultaneous localization and mapping problem~\cite{stachniss2016handbook-slamchapter} and additionally adding relevant information for the driving domain. Levinson \etal~\cite{levinson2007rss} utilize GPS, IMU, and LiDAR data to build HD maps for localization. They generate a 2D surface image of ground reflectivity in the infrared spectrum and define an observation model that uses these intensities. The uncertainty in intensity values has been handled by building a prior map~\cite{levinson2010icra, wolcott2015icra}.  
Barsan \etal~\cite{barsan2018corl} use a fully convolutional neural network~(CNN) to perform online-to-map matching for improving the robustness to dynamic objects and eliminating the need for LiDAR intensity calibration. Their approach shows a strong performance but requires a good GPS prior for operation.  
Based on this approach, Wei \etal~\cite{wei2019cvpr} proposed a learning-based compression method for HD maps. Merfels and Stachniss~\cite{merfels2016iros} present an efficient chain graph-like pose-graph for vehicle localization exploiting graph optimization techniques and different sensing modalities. Based on this work, Wilbers et al.~\cite{wilbers2019icra} propose a LiDAR-based localization system performing a combination of local data association between laser scans and HD map features, temporal data association smoothing, and a map matching approach for robustification.

Other approaches aim at performing LiDAR-based place recognition to initialize localization. For example, Kim \etal~\cite{kim2019ral} transform point clouds into scan context images and train a CNN based on such images. They generate scan context images for both the current frame and all grid cells of the map and compare them to estimate the current location as the cell presenting the largest score. 
Yin \etal~\cite{yin2019tits} propose a Siamese network to first generate fingerprints for LiDAR-based place recognition and then use iterative closest points to estimate the metric poses. 
Cop~\etal~\cite{cop2018icra} propose a descriptor for LiDAR scans based on intensity information. Using this descriptor, they first perform place recognition to find a coarse location of the robot, eliminate inconsistent matches using RANSAC, and then refine the estimated transformation using iterative closest points.
In contrast to approaches that perform place recognition first, our approach integrates a neural network-based observation model into an MCL framework to estimate the robot pose.  

Recently, several approaches exploiting semantic information for 3D LiDAR localization have been proposed. 
In our earlier work~\cite{chen2017ssrr}, we used a camera and a LiDAR to detect victims and localize the robot in an urban search and rescue environment. 
Ma \etal~\cite{ma2019iros} combine semantic information such as lanes and traffic signs in a Bayesian filtering framework to achieve accurate and robust localization within sparse HD maps.  
Yan \etal~\cite{yan2019ecmr} exploit buildings and intersections information from a LiDAR-based semantic segmentation system~\cite{milioto2019iros} to localize on OpenStreetMap. 
Schaefer \etal~\cite{schaefer2019ecmr} detect and extract pole landmarks from 3D LiDAR scans for long-term urban vehicle localization. 
Tinchev \etal~\cite{tinchev2019ral} propose a learning-based method to match segments of trees and localize in both urban and natural environments.
Dub{\'e} \etal~\cite{dube2018ral} propose to perform localization by extracting segments from 3D point clouds and matching them to accumulated data.
Whereas, Zhang \etal~\cite{zhang2018iros} utilize both ground reflectivity features and vertical features for localizing autonomous car in rainy conditions.
In our previous work~\cite{chen2019iros}, we also exploit semantic information~\cite{milioto2019iros, behley2019iccv} to improve the localization and mapping results by detecting and removing dynamic objects.

Different to the above discussed methods~\cite{barsan2018corl, ma2019iros, wei2019cvpr}, which use GPS as prior for localization, our method only exploits LiDAR information to achieve global localization without using any GPS information. Moreover, our approach uses range scans without explicitly exploiting semantics or extracting landmarks. 
Instead, we rely on CNNs to predict the overlap between range scans and their yaw angle offset and use this information as an observation model for Monte-Carlo localization. The localization approach proposed in this paper is based on our previous work called OverlapNet~\cite{chen2020rss}, which focuses on loop-closing for 3D LiDAR-based SLAM.

%%%%%%%%%%%%%%%%%%%%%%%%%%%%%%%%%%%%%%%%%%%%%%%%%%%%%%%%%%%%%%%%%%%%%%%%%%%%%%%%
\section{Our Approach}
\label{sec:main}

The key idea of our approach is to exploit the neural network, OverlapNet~\cite{chen2020rss} that can estimate the overlap and yaw angle offset of two scans for building an observation model for localization in a given map.
To this end, we first generate virtual scans at 2D locations on a grid rendered from the aggregated point cloud of the map.
We train the network completely self-supervised on the map used for localization~(see \secref{sec:approach-overlapnet}).
We compute features using our pre-trained network that allows us to compute the overlap and yaw angles between a query and virtual scans~(see \secref{sec:approach-map}).
Finally, we integrate an observation model using the overlap~(see \secref{sec:approach-sensor-overlap}) and a separate observation model for the yaw angle estimates~(see \secref{sec:approach-sensor-yaw}) in a particle filter to perform localization~(see \secref{sec:approach-mcl}).

%%%%%%%%%%%%%%%%%%%%%%%%%%%%%%%%%%%%%%%%
\subsection{OverlapNet}
\label{sec:approach-overlapnet}

We proposed the so-called OverlapNet to detect loop closures candidates for a 3D LiDAR-based SLAM. The idea of overlap has its origin in the photogrammetry and computer vision community~\cite{hussain2004manual}. The intuition is that to successfully match two images and calculate their relative pose, the images must overlap.
This can be quantified by defining the overlap percentage as the percentage of pixels
in the first image, which can successfully be projected back into the second
image. 

In OverlapNet, we use the idea of overlap for range images generated from 3D LiDAR scans exploiting the range information explicitly. First, we generate from the LiDAR scans a range-image like input tensor~$\mathbf{I} \in \RR^{H \times W \times 4}$, where each pixel~$(i,j)$ in the range-image like representation corresponds to the depth and the corresponding normal, \ie, $\mathbf{I}(i,j) = (\vec{r}, \vec{n}_x,  \vec{n}_y, \vec{n}_z)$, where $\vec{r}= ||\vec{p}||_2$.
The indices for inserting points and estimated normals are computed using a spherical projection that maps points~$\vec{p} \in \RR^3$ to two-dimensional coordinates $(i,j)$, also see \cite{chen2020rss, chen2019iros}.

OverlapNet uses a siamese network structure and takes the tensors~$\mat{I}_1$ and~$\mat{I}_2$ of two scans as input for the legs to generate feature volumes, $\mat{F}_1$ and $\mat{F}_2$.
These two feature volumes are then used as inputs for two heads. 
One head~$\set{H}_{\text{overlap}}(\mat{F}_1, \mat{F}_2)$ estimates the overlap percentage of the scans and the other head~$\set{H}_{\text{yaw}}(\mat{F}_1, \mat{F}_2)$ estimates the relative yaw angle. 

For training, we determine ground truth overlap and yaw angle values using known poses estimated by a SLAM approach~\cite{behley2018rss}.
Given these poses, we can train OverlapNet completely self-supervised.
For more details on the network structure and the training procedure, we are referring to our prior work \cite{chen2020rss}.

%%%%%%%%%%%%%%%%%%%%%%%%%%%%%%%%%%%%%%%%
\subsection{Map of Virtual Scans}  % revised according to reviews
\label{sec:approach-map}

OverlapNet requires two LiDAR scans as input. One is the current scan and the second one has to be generated from the map. Thus, we build a map of virtual LiDAR scans given an aggregated point cloud by using a grid of locations with grid resolution $\gamma$, where we generate virtual LiDAR scans for each location. The grid resolution is a trade-off between the accuracy and storage size of the map. 
Instead of storing these virtual scans, we just need to use one leg of the OverlapNet to obtain a feature volume~$\mat{F}$ using the input tensor~$\mat{I}$ of this virtual scan.
Storing the feature volume instead of the complete scan has two key advantages: (1) it uses more than an order less space than the original point cloud (ours: $0.1\,$Gb/km, raw scans: $1.7\,$Gb/km)  and (2) we do not need to compute the~$\mat{F}$ during localization on the map.
The features volumes of the virtual scans are then used to compute overlap and yaw angle estimates with a query scan that is the currently observed LiDAR point cloud in our localization framework.

%%%%%%%%%%%%%%%%%%%%%%%%%%%%%%%%%%%%%%%%
\subsection{Monte-Carlo Localization}
\label{sec:approach-mcl}

Monte-Carlo localization (MCL) is a localization algorithm based on the particle filter proposed by Dellaert \etal~\cite{dellaert1999icra}. 
Each particle represents a hypothesis for the robot's or autonomous vehicle's 2D pose $\vec{x}_t = (x, y, \theta)_t$ at time $t$. 
When the robot moves, the pose of each particle is updated with a prediction based on a motion model with the control input $\vec{u}_t$. 
The expected observation from the predicted pose of each particle is then compared to the actual observation $\vec{z}_t$ acquired by the robot to update the particle's weight based on an observation model. 
Particles are resampled according to their weight distribution and resampling is triggered  whenever the effective number of particles, see for example~\cite{grisetti2007tro}, drops below 50\% of the sample size.
After several iterations of this procedure, the particles will converge around the true pose.

MCL realizes a recursive Bayesian filtering scheme. The key idea of this approach is to maintain a probability density~$p(\vec{x}_t\mid\vec{z}_{1:t},\vec{u}_{1:t})$ of the pose~$\vec{x_t}$ at time~$t$ given all observations~$\vec{z}_{1:t}$ up to time~$t$ and motion control inputs~$\vec{u}_{1:t}$ up to time~$t$. This posterior is updated as follows:
\begin{align}
  &p(\vec{x}_t\mid\vec{z}_{1:t},\vec{u}_{1:t}) = \eta~p(\vec{z}_t\mid\vec{x}_{t}) \cdot\nonumber\\
  &\;\;\int{p(\vec{x}_t\mid\vec{u}_{t}, \vec{x}_{t-1})~p(\vec{x}_{t-1} \mid \vec{z}_{1:t-1},\vec{u}_{1:t-1})\ d\vec{x}_{t-1}},
\label{eq:bayesian}
\end{align}
where~$\eta$ is a normalization constant, $p(\vec{x}_t\mid\vec{u}_{t}, \vec{x}_{t-1})$~is the motion model, and $p(\vec{z}_t\mid\vec{x}_{t})$~is the observation model.

This paper focuses on the observation model. For the motion model, we follow a standard odometry model for vehicles~\cite{thrun2005probrobbook}.

We split the observation model into two parts:
\begin{align}
  p(\vec{z}_t\mid\vec{x}_{t}) &= p_{L} \left(\vec{z}_{t} \mid \vec{x}_{t} \right)~p_{O} \left( \vec{z}_{t} \mid \vec{x}_{t} \right) ,
%\label{eq:sensor}
\end{align}
where~$\vec{z}_{t}$ is the LiDAR observation at time~$t$, $p_{L} \left(\vec{z}_{t} \mid \vec{x}_{t} \right)$ is the probability encoding the location $(x, y)$ agreement between the current query LiDAR scan and the virtual scan at the grid where the particle locate and $p_{O} \left( \vec{z}_{t} \mid \vec{x}_{t} \right)$ is the probability encoding the yaw angle $\theta$ agreement between the same pairs of scans.

%%%%%%%%%%%%%%%%%%%%%%%%%%%%%%%%%%%%%%%%
\subsection{Overlap Observation Model}
\label{sec:approach-sensor-overlap}

\begin{figure}[t]
  \vspace{0.15cm}
  \centering
  \includegraphics[width=0.322\linewidth]{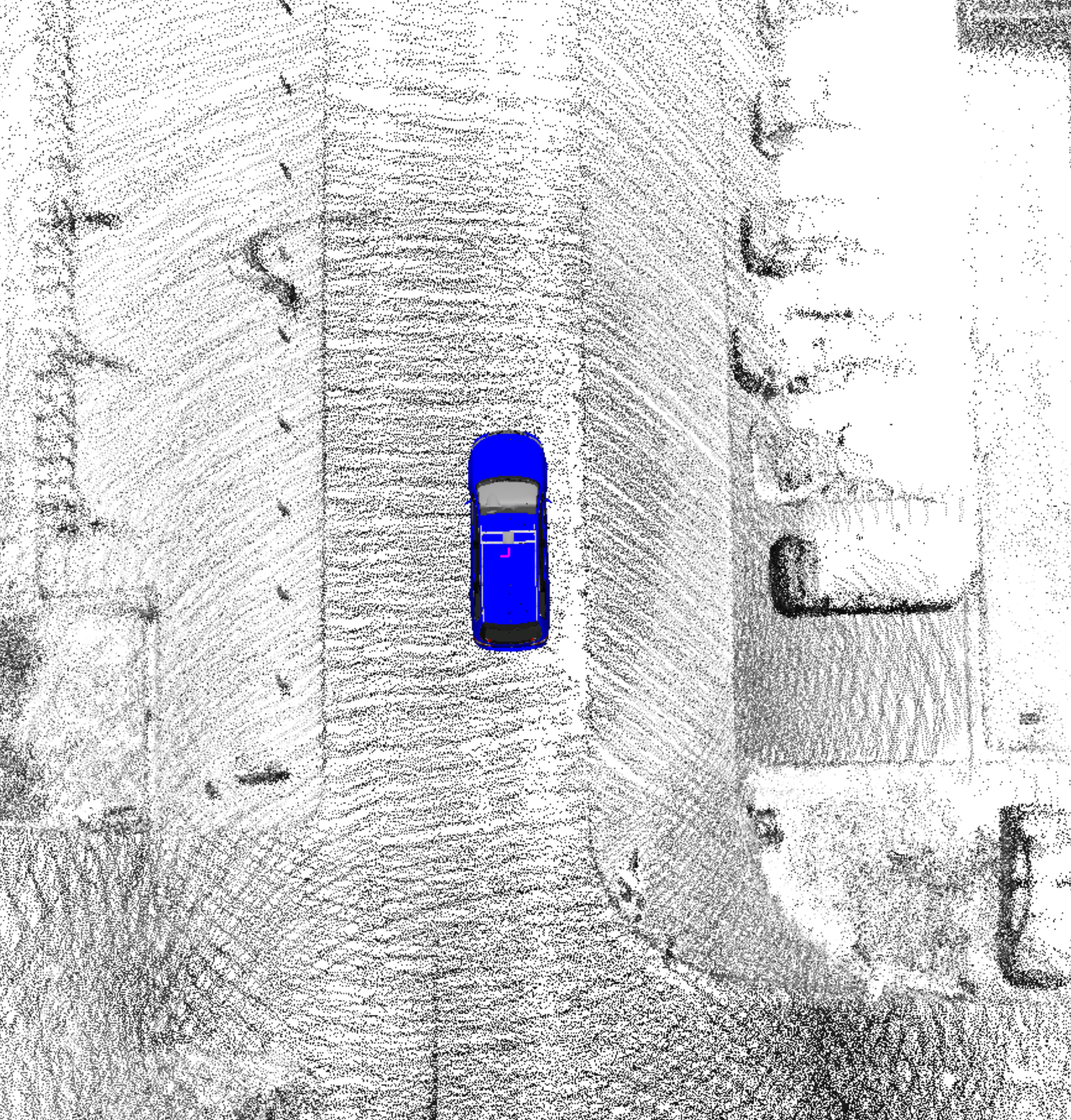}
  \includegraphics[width=0.322\linewidth]{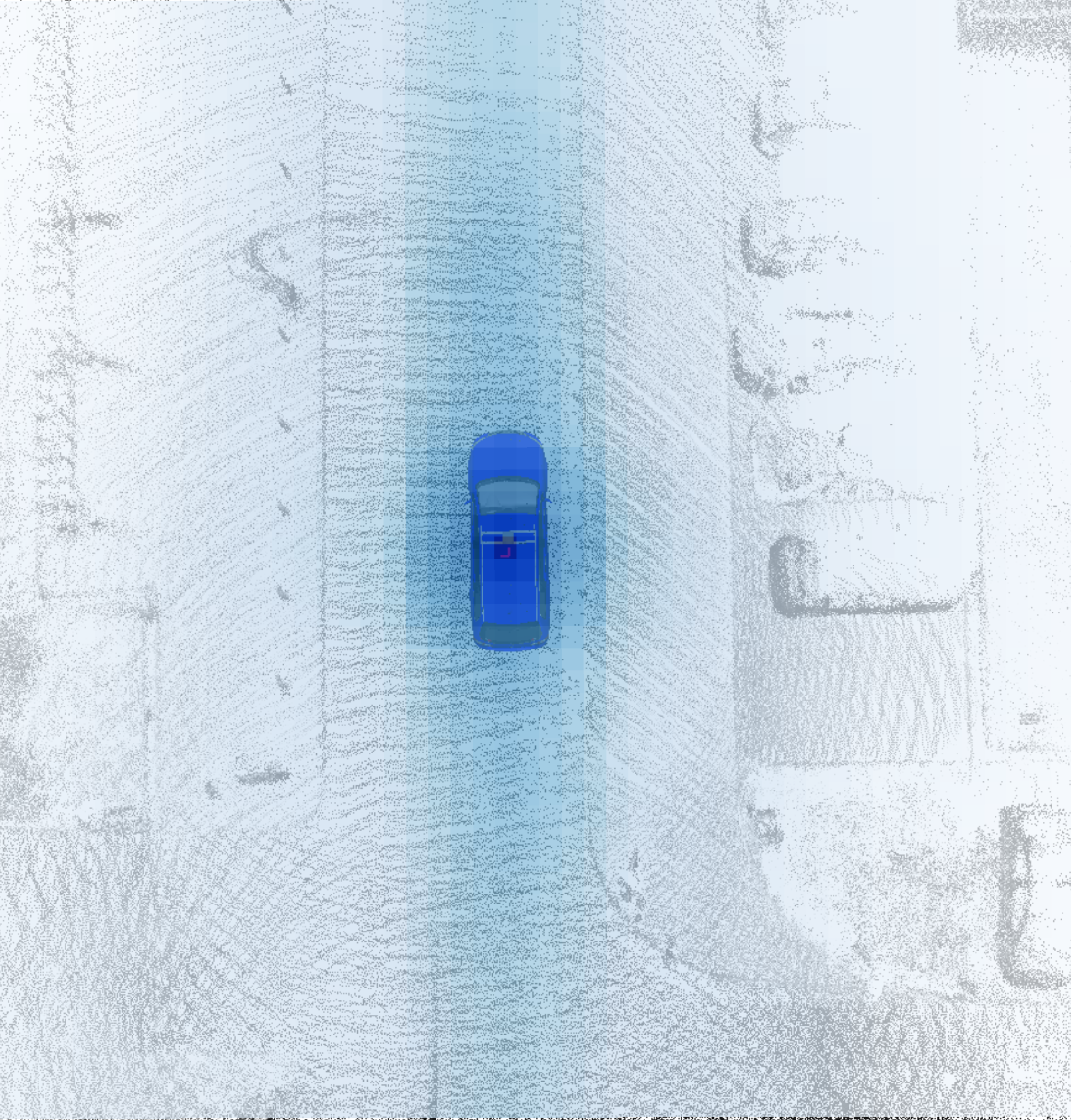}
  \includegraphics[width=0.325\linewidth]{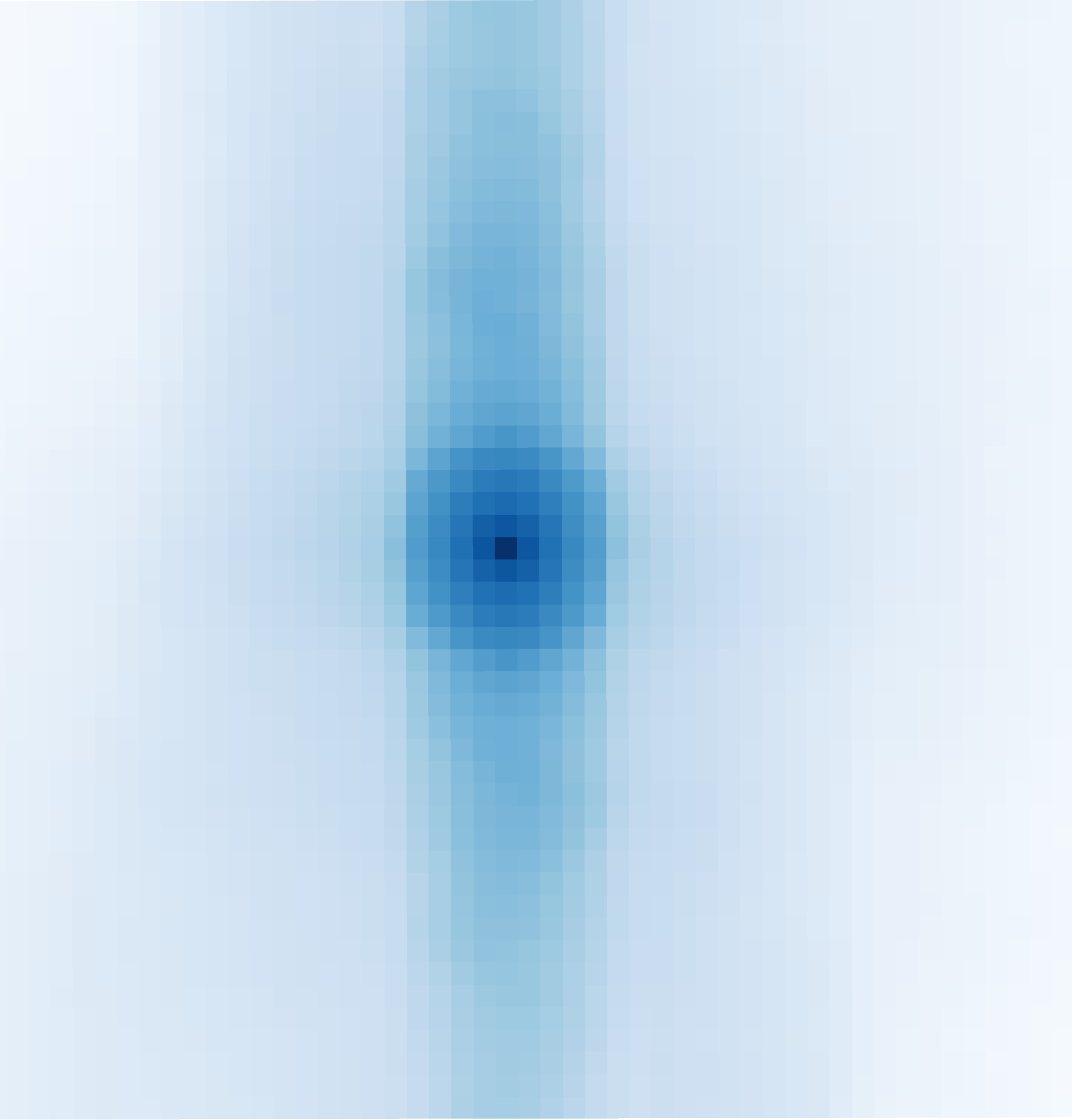}
  \caption{Overlap observation model. Local heatmap of the scan at the car's position with respect to the map. Darker blue shades correspond to higher probabilities.}
  \label{fig:overlap_sensor}
  \vspace{-0.3cm}
\end{figure}

Given a particle $i$ with the state vector~$(x_i, y_i, \theta_i)$, the overlap estimates encode the location agreement between the query LiDAR scan and virtual scans of the grid cells where particles locate. 
It can be directly used as the probability:
\begin{align}
  p_{L} \left(\vec{z}_{t} \mid \vec{x}_{t} \right) &\propto f\left(\vec{z}_{t}, \vec{z}_{i}; \vec{w} \right),
%\label{eq:sensor}
\end{align}
where~$f$ corresponds to the neural network providing the overlap estimation between the input scans $\vec{z}_{t}, \vec{z}_{i}$ and $\vec{w}$ is the pre-trained weights of the network.
$\vec{z}_{t}$ and $\vec{z}_{i}$ are the current query scan and a virtual scan of one $(x, y)$ location respectively. Note that no additional hyperparameter is needed to formulate our observation model for localization.

For illustration purposes, \figref{fig:overlap_sensor} shows the probabilities of all grid cells in a local area calculated by the overlap observation model. The blue car in the figure shows the current location of the car. The probabilities calculated by the overlap observation model can represent well the hypotheses of the current location of the car.

Typically, a large number of particles are used, especially when the environment is large. However, a large amount of particles will increase the computation time linearly. When applying the overlap observation model, particles could still obtain relatively large weights as long as they are close to the actual pose, even if not in the exact same position. This allows us to use fewer particles to achieve a high success rate of global localization. 

Furthermore, the overlap estimation only encodes the location hypotheses. Therefore, if multiple particles locate in the same grid area, only a single inference against the nearest virtual scan of the map needs to be done, which can further reduce the computation time.

%%%%%%%%%%%%%%%%%%%%%%%%%%%%%%%%%%%%%%%%
\subsection{Yaw Angle Observation Model}
\label{sec:approach-sensor-yaw}

Given a particle $i$ with the state vector~$(x_i, y_i, \theta_i)$, the yaw angle estimates encode the orientation agreement between the query LiDAR scan and virtual scans of the corresponding grids where particles locate. We formulate the orientation probability as follows:
\begin{align}
  p_{O} \left(\vec{z}_{t} \mid \vec{x}_{t} \right) &\propto \exp{\left( -\frac{1}{2} \frac{\Big(g\left(\vec{z}_{t}, \vec{z}_{i}; \vec{w} \right) - \theta_i \Big)^2}{\sigma^2_\theta}\right)} ,
\label{eq:yaw_sensor}
\end{align}
where~$g$ corresponds to the neural network providing the yaw angle estimation between the input scans $\vec{z}_{t}, \vec{z}_{i}$ and $\vec{w}$ is the pre-trained weights of the network.
$\vec{z}_{t}$ and $\vec{z}_{i}$ are the current query scan and a virtual scan of one particle respectively.

When generating the virtual scans of the grid map, all virtual scans will be set facing the absolute 0\si{\degree} yaw angle direction. By doing this, the estimated relative yaw angle between the query scan and the virtual scan indicates the absolute yaw angle of the current query scan. \eqref{eq:yaw_sensor} assumes a Gaussian measurement error in the heading. 

By combining overlap and yaw angle estimation, the proposed observation model will correct the weights of particles considering agreements between the query scan and the map with the full pose~$(x, y, \theta)$.  

%%%%%%%%%%%%%%%%%%%%%%%%%%%%%%%%%%%%%%%%%%%%%%%%%%%%%%%%%%%%%%%%%%%%%%%%%%%%%%%%
\section{Experimental Evaluation}
\label{sec:exp}

In this paper, we use a grid representation and generate virtual $360\si{\degree}$ scans for each grid point. 
The resolution of the grid is $\gamma=20\,$~cm. 
When generating the virtual scans, we set the yaw angle to $0\si{\degree}$. Therefore, when estimating the relative yaw angle between the query frame and the virtual frames, the result will indicate the absolute yaw of the query frame. For the yaw angle observation model in~\eqref{eq:yaw_sensor}, we set $\sigma_{\theta} = 5\si{\degree}$.  
To achieve global localization, we train a new model only based on the map scans and the generated virtual scans. 

The main focus of this work is a new observation model for LiDAR-based localization. 
Therefore, when comparing different methods, we only change the observation model $p(\vec{z}_t|\vec{x}_t)$ of the MCL framework and keep the particle filter-based localization the same. The motion model is the typical odometry model~\cite{thrun2005probrobbook}.

%%%%%%%%%%%%%%%%%%%%%%%%%%%%%%%%%%%%%%%%

\subsection{Car Dataset}
\begin{figure}[t]
  \vspace{0.15cm}
  \centering
  \includegraphics[width=0.95\linewidth]{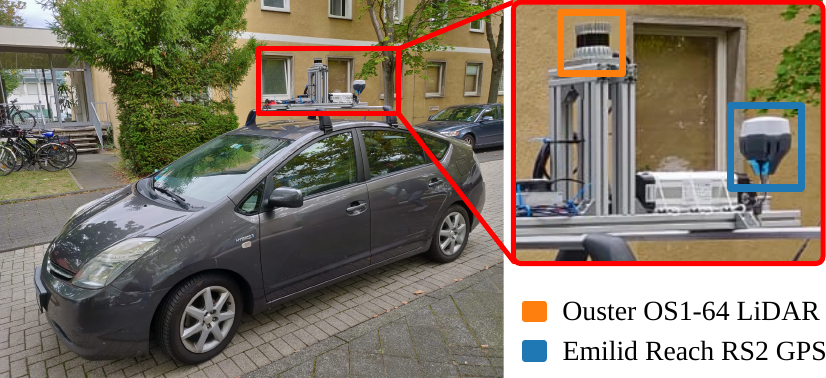}
  \caption{Sensor setup used for data recording: Ouster OS1-64 LiDAR sensor plus GNSS information from a Emilid Reach RS2.}
  \label{fig:ipb-car}
\end{figure}

The dataset used in this paper was collected using a self-developed sensor platform illustrated in~\figref{fig:ipb-car}.
To test LiDAR-based global localization, a large-scale dataset has been collected in different seasons with multiple sequences repeatedly exploring the same crowded urban area. 
For our car dataset, we performed a 3D LiDAR SLAM \cite{behley2018rss} combined with GPS information to create near ground truth poses. 
During localization, we only use LiDAR scans for global localization without using GPS. 

The dataset has three sequences that were collected at different times of the year, sequence~00 in September 2019, sequence ~01 in November 2019, and sequence~02 in February 2020. 
The whole dataset covers a distance of over $10$~km.  
We use LiDAR scans from sequence~02 to build the virtual scans and use sequence~00 and~01 for localization. 
As can be seen from~\figref{fig:motivation}, the appearance of the environment changes quite significantly since the dataset was collected in different seasons and in crowded urban environments, including changes in vegetation, but also cars at different locations and moving people.

%%%%%%%%%%%%%%%%%%%%%%%%%%%%%%%%%%%%%%%%
\subsection{Different Observation Models}

\begin{figure}[t]
  \centering
  \subfigure[Corresponding submap]{\includegraphics[width=0.45\linewidth]{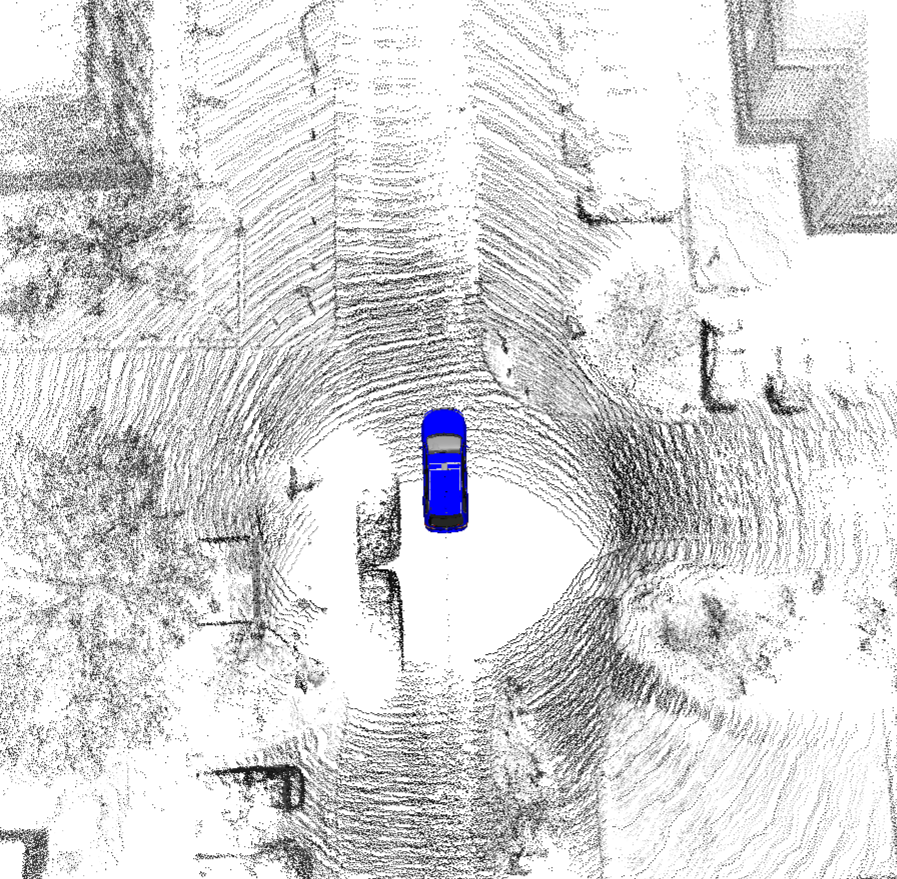}}
  \subfigure[Overlap-based $p(z|x)$]{\includegraphics[width=0.45\linewidth]{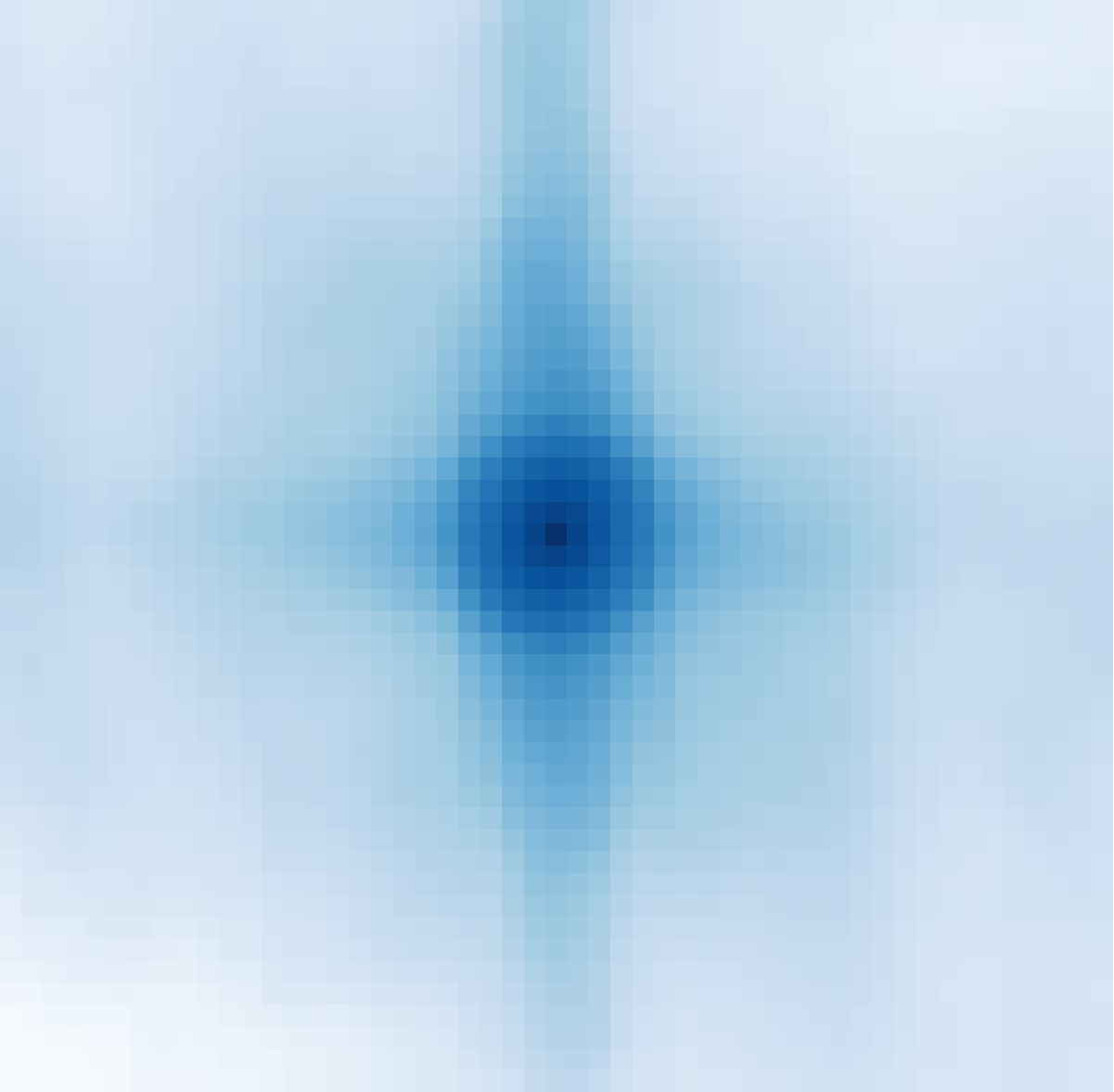}}
  \subfigure[Histogram-based $p(z|x)$]{\includegraphics[width=0.45\linewidth]{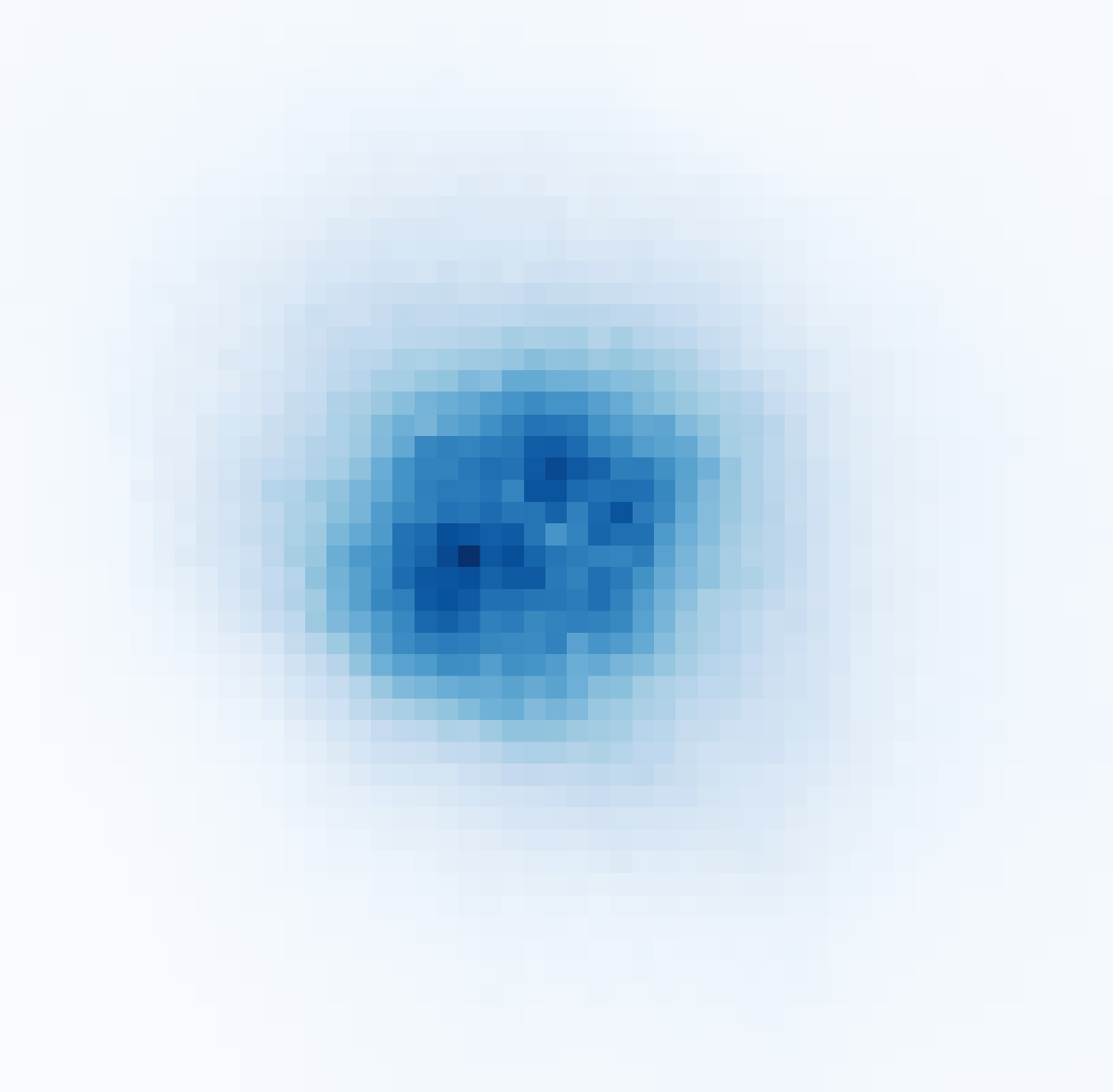}}
  \subfigure[Beam-end $p(z|x)$]{\includegraphics[width=0.45\linewidth]{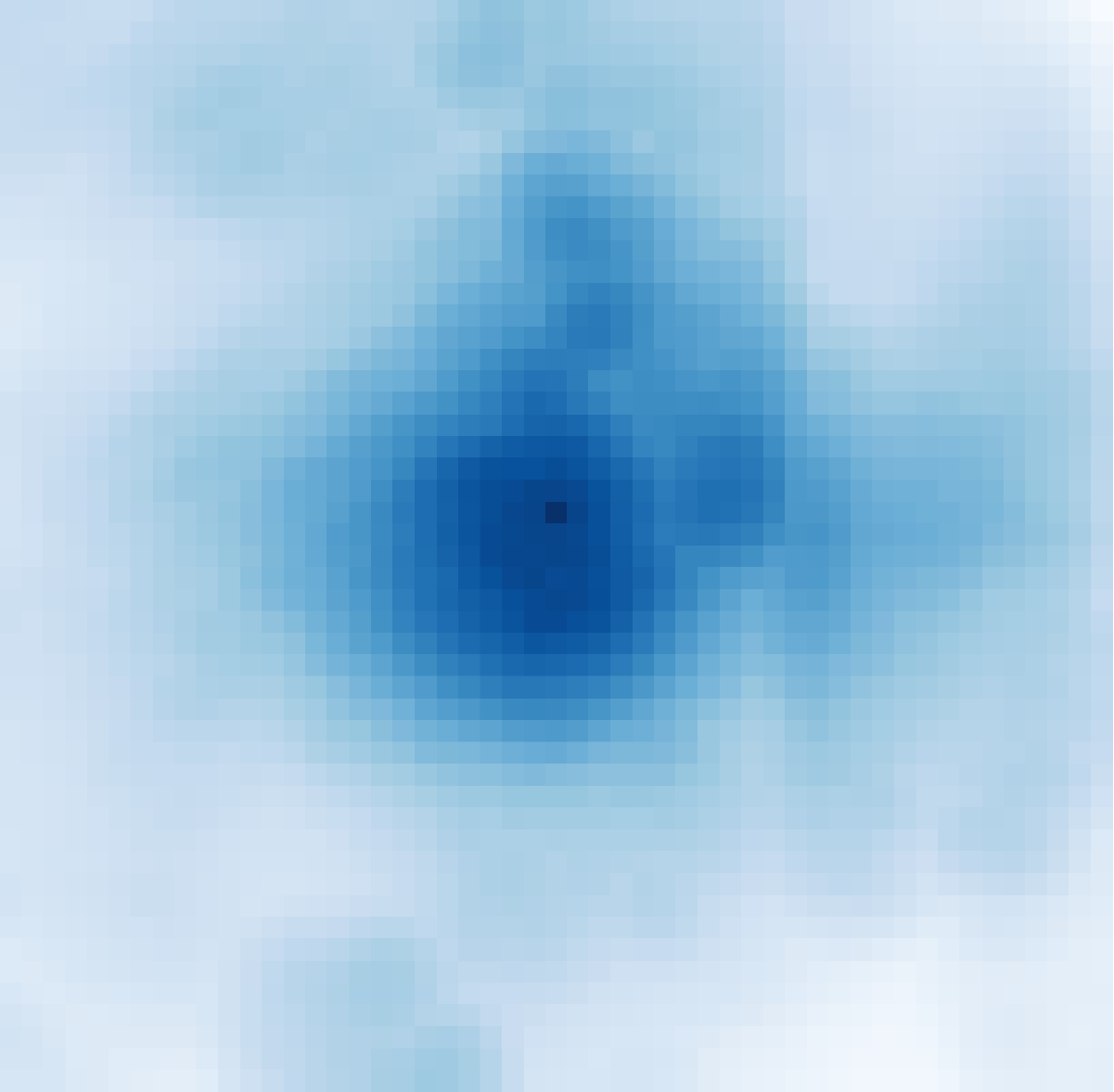}}
  \caption{Heatmaps of different observation models used in the experiments generated for the same query scan and map.}
  \label{fig:sensor_models}
\end{figure}

In the following experiments, we use the same MCL framework and only exchange the observation models. 
We compare our observation model with two baseline observation models: the typical beam-end model~\cite{thrun2005probrobbook} and a histogram-based model derived from the work of R\"ohling \etal~\cite{roehling2015iros}. 

The beam-end observation model is often used for 2D LiDAR data.
For 3D LiDAR scans, it needs much more particles to make sure that it converges to the correct pose, which causes the computation time to increase substantially. In this paper, we implement the beam-end model with a down-sampled point cloud map using voxelization with a resolution of $10$\,cm. 

Our second baseline for comparison is inspired by R\"ohling \etal~\cite{roehling2015iros}, which proposed a fast method to detect loop closures through the use of similarity measures on histograms extracted from 3D LiDAR data. 
The histogram contains the range information. We use a similar idea, but integrate it into the MCL framework as a baseline observation model. We employ the same grid map and virtual frames as used for our method with the histogram-based observation model. When updating the weights of particles, we will first generate the histograms of the current query scan and the virtual scans of grids where the particles locate. Then, we use the same Wasserstein distance to measure the similarity between them and update the weights of particles as follows: 
\begin{align}
 p(\vec{z}_t|\vec{x}_{t}) \propto d \left( h(\vec{z}_{t}), h(\vec{z}_{i} \right)),
\end{align}
where $d$ represents the Wasserstein distance between histograms $h(\vec{z}_{t}), h(\vec{z}_{i})$ of LiDAR scan $\vec{z}_{t}, \vec{z}_{i}$. 

\figref{fig:sensor_models} shows the 2D heatmaps in $x$ and $y$ calculated for the different observation models. As can be seen, the proposed observation model tends to give higher weights to the positions along the road, which leads to a higher success rate when the vehicle aims to localize in an urban environment. We will show the numerical results which verify that our method can achieve a high success rate with much fewer particles in the next sections.

\begin{figure}[t]
  \centering
  \vspace{0.2cm}
  \subfigure[Trajectory from sequence 00]{\includegraphics[width=0.75\linewidth]{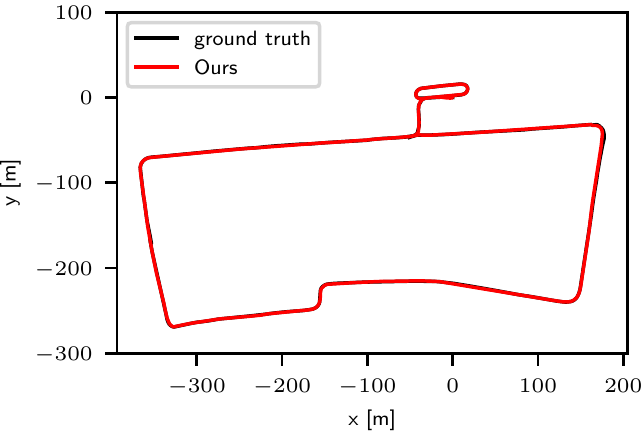}}
  \subfigure[Trajectory from sequence 01]{\includegraphics[width=0.75\linewidth]{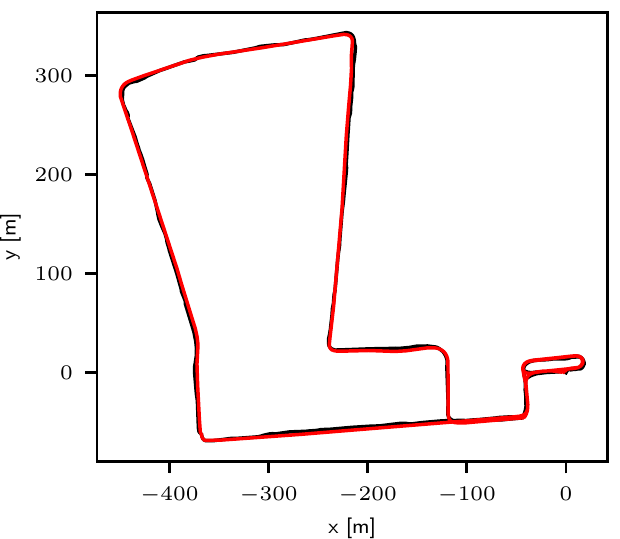}}
  \caption{Localization results of our method with $10,000$ particles on two sequences recorded with the setup depicted in \figref{fig:ipb-car}. Shown are the ground truth trajectory (black) and our estimated trajectory using our observation model (red).}
  \label{fig:loc_results}
\end{figure}

%%%%%%%%%%%%%%%%%%%%%%%%
\subsection{Localization Performance}
\label{sec:loc_results}

The experiment presented in this section is designed to show the performance of our approach and to support the claim that it is well suited for global localization.

First of all, we show the general localization results tested with two sequences in~\figref{fig:loc_results}. The qualitative results show that, after applying our sensor-model, the proposed method can well localize in the map with only LiDAR data collected in highly dynamic environments at different times.

\begin{figure}[t]
  \vspace{0.2cm}
  \centering
  \subfigure[sequence 00]{\includegraphics[width=0.49\linewidth]{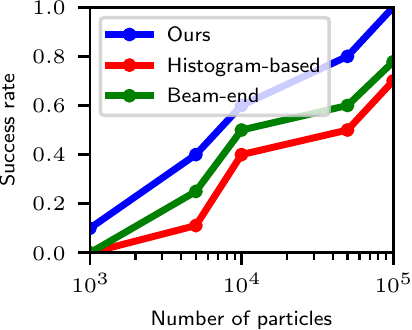}}
  \subfigure[sequence 01]{\includegraphics[width=0.49\linewidth]{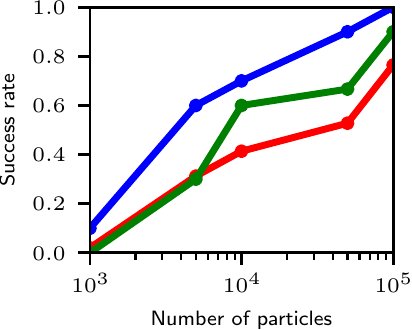}}
  \caption{Success rate of the different observation models for $10$ runs with~$N=$~\{$1,000$\,; $5,000$\,; $10,000$\,; $50,000$\,; $100,000$\} particles. Here, we use sequence 00 and sequence 01 to localize in sequence 02. We count runs as success if they converge to the ground truth location within $5\,$m.}
  \label{fig:success_rate}
%\end{figure}
  \vspace{1cm}
%\begin{figure}[t]
%\vspace{0.15cm}
  \centering
  \includegraphics[width=0.95\linewidth]{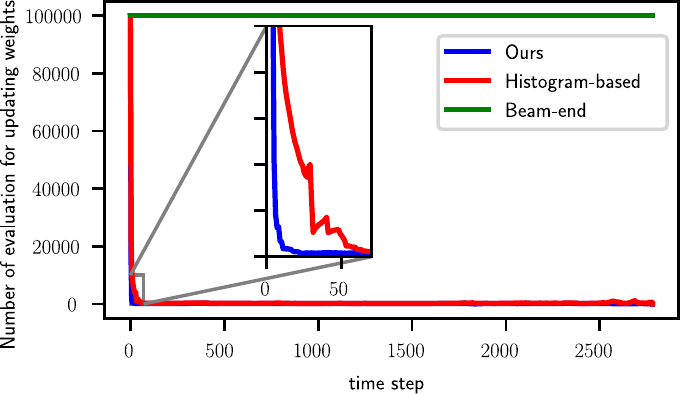}
  \caption{Number of observation model evaluations for updating the weights at each timestep with $100,000$ particles. The beam end model needs to be evaluated for each and every particle individually. The histogram-based method is more computationally efficient, while our proposed method still needs the fewest evaluations.}
  \label{fig:computing_times}
\end{figure}

For quantitative results, we calculate the success rate for different methods with different particle numbers comparing our approach to different methods, as shown in~\figref{fig:success_rate}. The x-axis represents the number of particles used during localization, while the y-axis is the success rate of different setups. The success rate for a specific setup of one method is calculated using the number of success cases divided by the total numbers of the tests. To decide whether one test is successful or not, we check the location error by every 100~frames after converging. If the location error is smaller than a certain threshold, we count this run as a success case.

We test our method together with two baselines using five different numbers of particles~$N=$~\{$1,000$\,; $5,000$\,; $10,000$\,; $50,000$\,; $100,000$\}. For each setup, we sample 10~trajectories and perform global localization. 

Quantitative results of localization accuracy are shown in~\tabref{tab:location_results} and~\tabref{tab:yaw_results}.
\tabref{tab:location_results} shows the location error of all methods tested with both sequences. The location error is defined as the root mean square error~(RMSE) of each test in terms of $(x,\,y)$ Euclidean error with respect to the ground truth poses. \tabref{tab:location_results} shows the mean and the standard deviation of the error for each observation model. Note that the location error is only calculated for success cases. 

\tabref{tab:yaw_results} shows the yaw angle error. It is the RMSE of each test in terms of yaw angle error with respect to the ground truth poses. The table shows the mean and the standard deviation of the error for each observation model. As before, the yaw angle error is also only calculated for cases in which the global localization converged.

% results on road set
\begin{table}[t]
\centering
\vspace{0.5cm}
\caption{Location results}
\scalebox{1}{
\scriptsize{
\begin{tabular}{cccccccc}
\toprule
Sequence & \multicolumn{3}{c}{Location error [meter]} \\
\cline{2-4}
 & Beam-end & Histogram-based & Ours \Tstrut \\
\midrule
0 & $0.92$~$\pm$~$0.27$ & $1.85$~$\pm$~$0.34$ & $\mathbf{0.81 \pm 0.13}$ \\
1 & $\mathbf{0.67 \pm 0.11}$ & $1.86$~$\pm$~$0.34$ & $0.88$~$\pm$~$0.07$ \\
\midrule
\end{tabular}
}
}
\label{tab:location_results}
%\vspace{-0.4cm}
\end{table} 

% results on road set
\begin{table}[t]
\centering
\vspace{0.2cm}
\caption{Yaw angle results}
\scalebox{1}{
\scriptsize{
\begin{tabular}{ccccc}
\toprule
Sequence & \multicolumn{3}{c}{Yaw angle error [degree]} \\
\cline{2-4}
 & Beam-end & Histogram-based & Ours \Tstrut \\
\midrule
0 & $1.87$~$\pm$~$0.47$ & $3.10$~$\pm$~$3.07$ & $\mathbf{1.74 \pm 0.11}$ \\
1 & $2.10$~$\pm$~$0.59$ & $3.11$~$\pm$~$3.08$ & $\mathbf{1.88 \pm 0.09}$ \\
\midrule
\end{tabular}
}
}
\label{tab:yaw_results}
%\vspace{-0.4cm}
\end{table}

As can be seen from the results, our method achieves higher success rates with a smaller number of particles compared to the baseline methods, which also makes the proposed method faster than baseline methods.
Furthermore, our method converges already with $100,000$~particles in all cases, whereas the other observation models still need more particles to sufficiently cover the state space.
Moreover, the proposed method gets similar performance in location error comparing to the baseline methods but it achieves better results in yaw angle estimation. This is because the proposed method decouples the location and yaw angle estimation and, therefore, can exploit more constraint in yaw angle corrections. 

To sum up, the proposed method outperforms the baselines method in terms of success rate, while getting similar results in terms of location error. Moreover, our method outperforms baseline methods in yaw angle estimation, because of the proposed de-coupled observation model. Furthermore, our method is faster than the baseline method. The runtime details will be shown in the next experiment.

%%%%%%%%%%%%%%%%%%%%%%%%
\subsection{Runtime}

In this experiment, we show the \emph{number} of observation model evaluations necessary for updating the weights at each time step in~\figref{fig:computing_times}. This is a fairer way to compare the computational cost of different methods, since our neural network-based method uses GPU to concurrently updating the weights of particles, while the other methods only use CPU. As can be seen, our method needs a smaller number of observation model evaluations to update the weights for all particles. 
This is because we only need to perform the network inference for all particles which are localized in the same grid cell once. For one incoming frame and the virtual frame of that grid cell, the inputs of the network and thus the outputs are the same for all particles in that cell.

We tested our method on a system equipped with an Intel i7-8700 with 3.2 GHz and an Nvidia GeForce GTX 1080 Ti with 11~GB of memory. For initializing in the large-scale map, the worst case will take around $43\,$s to process one frame. However, after converging, the proposed method takes only 1~s on average to process one frame with $10,000$ particles.

%%%%%%%%%%%%%%%%%%%%%%%%%%%%%%%%%%%%%%%%%%%%%%%%%%%%%%%%%%%%%%%%%%%%%%%%%%%%%%%%
\section{Conclusion}
\label{sec:conclusion}

In this paper, we presented a novel observation model and integrated it into an MCL framework to solve the global localization problem.
Our method exploits OverlapNet to estimate the overlap and yaw angle between the current frame and the virtual frames generated at each particle using the pre-built map.
This allows us to successfully update the weights of particles with agreements of both location and orientation.
We implemented and evaluated our approach on an urban dataset collected with a car in different seasons and provided comparisons to other existing techniques. 
The experiments suggest that our approach can achieve a similar performance as other approaches in global localization while obtaining a higher success rate and lower computational time.

%%%%%%%%%%%%%%%%%%%%%%%%%%%%%%%%%%%%%%%%%%%%%%%%%%%%%%%%%%%%%%%%%%%%%%%%%%%%%%%%
%\emph{Future work: Use only if applicable -- but if so, use the following
%  sentence to start:}
In future work, we will test our method on more datasets with different types of LiDAR sensors.
We also plan to test our method with high definition  maps and want to exploit semantic information.

%%%%%%%%%%%%%%%%%%%%%%%%%%%%%%%%%%%%%%%%%%%%%%%%%%%%%%%%%%%%%%%%%%%%%%%%%%%%%%%%
% Only if applicable
%\section*{Acknowledgments}
%We thank XXX for fruitful discussions and for \dots

\bibliographystyle{plain}

% All new citations should go to new.bib. The file glorified.bib should go
% be the one from the ipb server. After paper or related work has been
% written merge the entries from new.bib to glorified.bib ON THE SERVER,
% replace the glorified.bib in this repository and empty the new.bib
\bibliography{glorified,new}

\end{document}